\pdfoutput=1
\documentclass[10pt, conference, compsocconf]{IEEEtran}

\usepackage{hyperref}
\usepackage{cleveref}
\usepackage{ifpdf}
\usepackage{cite}
\ifCLASSINFOpdf
   \usepackage[pdftex]{graphicx}
\fi
\usepackage[cmex10]{amsmath}
\usepackage[linesnumbered,ruled,vlined,commentsnumbered]{algorithm2e}
\usepackage[caption=false,font=footnotesize]{subfig}
\usepackage{amsfonts}
\usepackage{bm}
\usepackage{hyperref}
\makeatletter
\newcommand{\removelatexerror}{\let\@latex@error\@gobble}
\makeatother

\begin{document}
\def\IEEEbibitemsep{0pt}

\title{STAR: A Concise Deep Learning Framework for Citywide Human Mobility Prediction}
\author{
    \IEEEauthorblockN{Hongnian Wang and Han Su}
    \IEEEauthorblockA{School of Computer Science\\
                      Sichuan Normal University\\
                      Chengdu, China\\
                      e-mail: hongnianwang@gmail.com, jkxy\_sh@sicnu.edu.cn}
}

\maketitle

\begin{abstract}
Human mobility forecasting in a city is of utmost importance to transportation and public safety, but with the process of urbanization and the generation of big data, intensive computing and determination of mobility pattern have become challenging. This study focuses on how to improve the accuracy and efficiency of predicting citywide human mobility via a simpler solution. A spatio-temporal mobility event prediction framework based on a single fully-convolutional residual network (STAR) is proposed. STAR is a highly simple, general and effective method for learning a single tensor representing the mobility event. Residual learning is utilized for training the deep network to derive the detailed result for scenarios of citywide prediction. Extensive benchmark evaluation results on real-world data demonstrate that STAR outperforms state-of-the-art approaches in single- and multi-step prediction while utilizing fewer parameters and achieving higher efficiency.

\end{abstract}

\begin{IEEEkeywords}
convolutional neural network; residual learning; flow prediction; spatio-temporal data mining
\end{IEEEkeywords}

\section{Introduction}
The gathering of massive mobility event in urban area accelerates industrial reform, technological innovation, and lifestyle changes. Meanwhile, several problems, such as traffic jams, environmental deterioration, and increasing potential safety hazards, accompany the process of urbanization. This situation enables the prediction of human mobility events in urban environments, which is of importance to public safety, traffic management, and network optimization \cite{zheng2014urban}. For example, if managers are aware of the prediction results, they can quickly understand each region of public safety condition and apply precautionary measures in time. 

\begin{figure}[t!]
\centerline{\includegraphics[width=3.15in]{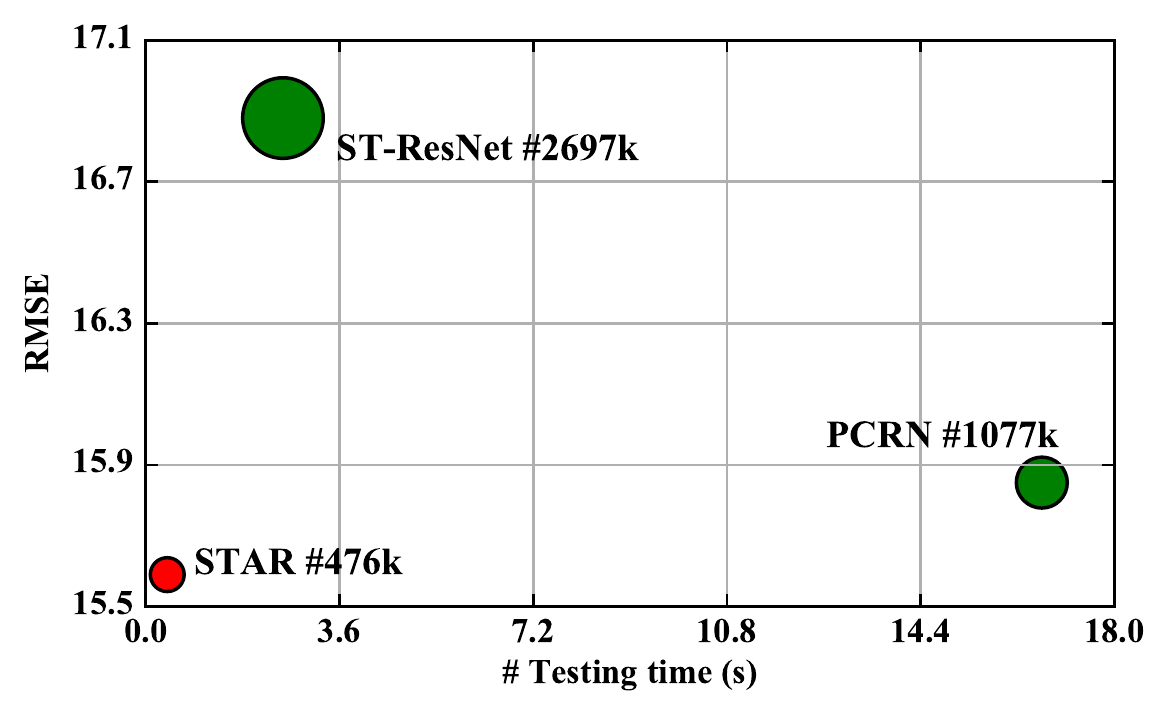}}
\caption{RMSE of recent DL models on TaxiBJ. \textit{(the red point is our model, which achieves lowest RMSE overall steps with fewer parameters and higher efficiency than state-of-the-art methods)}}
\label{fig:introduction}
\end{figure}

Human mobility is defined as the total number of humans passing through an area during a certain period. We can view this issue as a spatio-temporal (ST) prediction problem\cite{zheng2014urban}, which is challenging mainly because its patterns are affected by multiple and complex factors, including the spatial correlation among different regions, the temporal correlation among different time intervals and stochastic events, such as weather, accidents and so on \cite{zhang2017deep}. The shallow architecture cannot adequately describe these intricate pattern, and as the data increase, the performance deteriorates \cite{liu2018urban,lv2015traffic}.

\begin{figure}[!t]
    \centering
    \subfloat[]{\label{fig:st1}
    \includegraphics[width=0.78in]{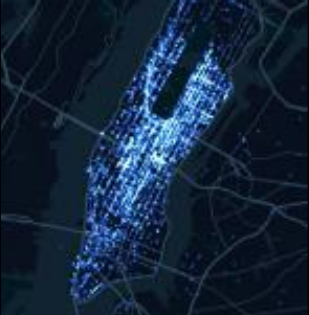}}
    \quad
    \quad
    \subfloat[]{\label{fig:st2}
    \includegraphics[width=0.78in]{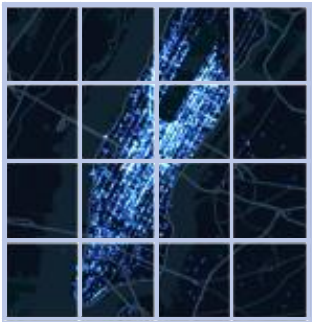}}
    \quad
    \quad
    \subfloat[]{\label{fig:st3}
    \includegraphics[width=0.78in]{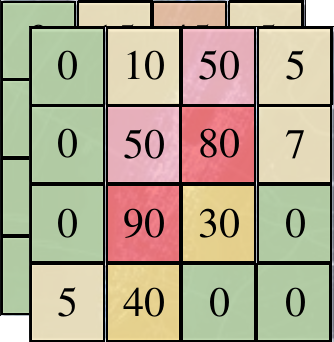}}\\
    \caption{The process of the grid-based method on NYC. \textit{(Using definition 1 to split a city \protect\subref{fig:st1} into $\bm{S}_t$ regions \protect\subref{fig:st2}, then count each region inflow and outflow $\bm{X}_t$ \protect\subref{fig:st3} by definition 2)}}
    \label{fig:stdata}
\end{figure}

Today, the large amount of available location-acquisition sensors, have resulted in large-scale, high-quality data with spatial coordinates and timestamps, which represent the mobility of moving objects, such as humans \cite{zheng2015trajectory}. Also, the data increase dramatically, which support citywide mobility flow forecasting, resulting in a highly challenging task. The deep learning (DL) model, a typical data-driven approach, can automatically identify the most representative features from a massive amount of mobility data, a task unable to fulfill with previous methods \cite{liu2018urban}. As a result of the rapid development of DL \cite{lecun2015deep}, employing DL models into big ST data prediction problems becomes a feasible task. To our best know, Zhang et al. \cite{zhang2017deep} firstly applied the grid-based method in this task, as shown in Fig.~\ref{fig:stdata}, which transformed historical trajectory data into image data, making it possible to employ a DL model to predict mobility events. Their method, termed ST-ResNet, showed state-of-the-art performance without any engineering features' requirement. To some extent, this is a new perspective for post-researchers, and several DL approaches have been proposed to predict ST data by adopting the grid-based method \cite{yuan2018hetero,chen2018exploiting,zonoozi2018periodic}, but these methods are computationally expensive, where the results are not significant enough at accuracy and efficiency.

ST-ResNet successfully introduced DL theory into the ST prediction problem, but the model has a large capacity and requires the updating of numerous parameters, making it hard to optimize and prone to over-learning. This is inconsistent with Occam's razor theory. Moreover, in practical applications, inaccurate model or delay prediction makes the measures meaningless or even results in additional losses. For instance, send traffic police officers to non-congested places. Applying the deep architecture model to real-world applications such as human mobility prediction problem still requires continued research.

Accordingly, we propose a simplified deep architecture, called STAR. Fig.~\ref{fig:introduction} shows the root mean squared (RMSE) performance of several recent DL models for mobility events prediction versus the number of parameters and testing time. STAR exhibits superior prediction performance with a highly concise network. The main novelties and contributions of STAR are as following:

\begin{itemize}
\item We propose a new idea to select historical data that can represent temporal correlation more reasonable than the previous methods. We prove that the learning power of convolution kernels in the channel dimension will be stronger than the previous fusion methods if we construct the input data reasonably.
\item STAR only employs a single network to model temporal correlation, thereby reducing the enormous number of parameters and improving the iteration speed of the model. To the best of our knowledge, our model is the first model to employ a single 2D convolution network for mobility event prediction.
\item We evaluate our framework in two representative real-world datasets. The experiments on single- and multi-step prediction show that STAR achieves the best performance with improved efficiency compared with state-of-the-art models, which proves that effective methods are usually simple and general.
\end{itemize}

\section{Preliminary}

\subsection{Notation}
\noindent {\bf Definition 1. Spatial attributes.} We split the target city into $I\times J$ grids based on the longitude and latitude and use $\bm{S}=(s^{0,0},...,s^{i,j})\in \mathbb{R}$ to represent spatial attributes of a city, where $s^{i,j}$ represents the $i$-th row and the $j$-th column area in the city.

\noindent {\bf Definition 2. Inflow/outflow.} According to the previous study \cite{zhang2017deep}, human mobility has two types of flows (\textit{i.e.}, inflow and outflow). The inflow and outflow of an area $s^{i,j}$ at the time interval $t$ are defined respectively as

\begin{equation}
  x_t^{in,i,j}=\sum_{Tr\in\mathbb{P}}|\{k>1|g_{k-1}\notin s^{i,j}\wedge{g_k}\in s^{i,j}\}|\label{eq1}
\end{equation}
\begin{equation}
  x_t^{out,i,j}=\sum_{Tr\in\mathbb{P}}|\{k>1|g_{k}\in s^{i,j}\wedge{g_{k+1}}\notin s^{i,j}\}|\label{eq2}
\end{equation}
where $Tr:g_1\to g_2\to ...\to g_{|Tr|}$ is a trajectory in $\mathbb{P}$, which is a collection of trajectories at the $t$-th time interval, and $g_k$ is the geospatial coordinate; $g_{k}\in s^{i,j}$ means the point $g_k$ lies within the $s^{i,j}$, and vice versa; $|\cdot|$ denotes the cardinality of a set. We use $\bm{X}_t\in \mathbb{R}^{2^t\times I \times J}$ to denote the historical observations, including all area of $ x_t^{in,i,j}$ and $ x_t^{out,i,j}$ in $\bm{S}$.

\noindent {\bf Problem statement.} Given the historical observations $\{\bm{X}_t|t=0,...,n-1\}$ to predict $\bm{X}_n$ in the future.

\subsection{Related DL models}

\begin{figure}[!t]
    \centering
    \subfloat[ST-ResNet]{\label{fig:model1}
    \includegraphics[height=1.9in]{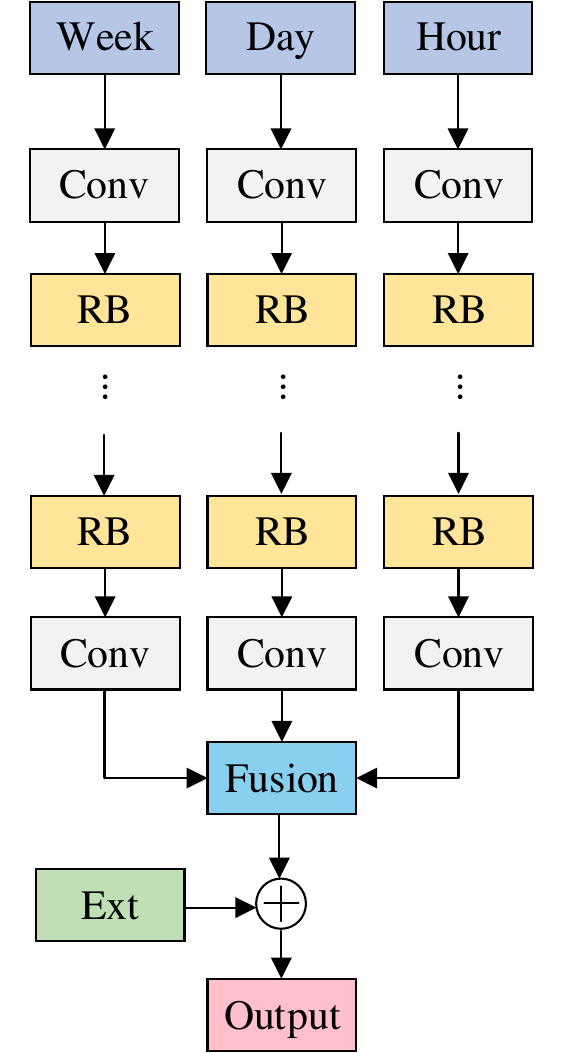}}
    \quad
    \subfloat[PCRN]{\label{fig:model2}
    \includegraphics[height=1.9in]{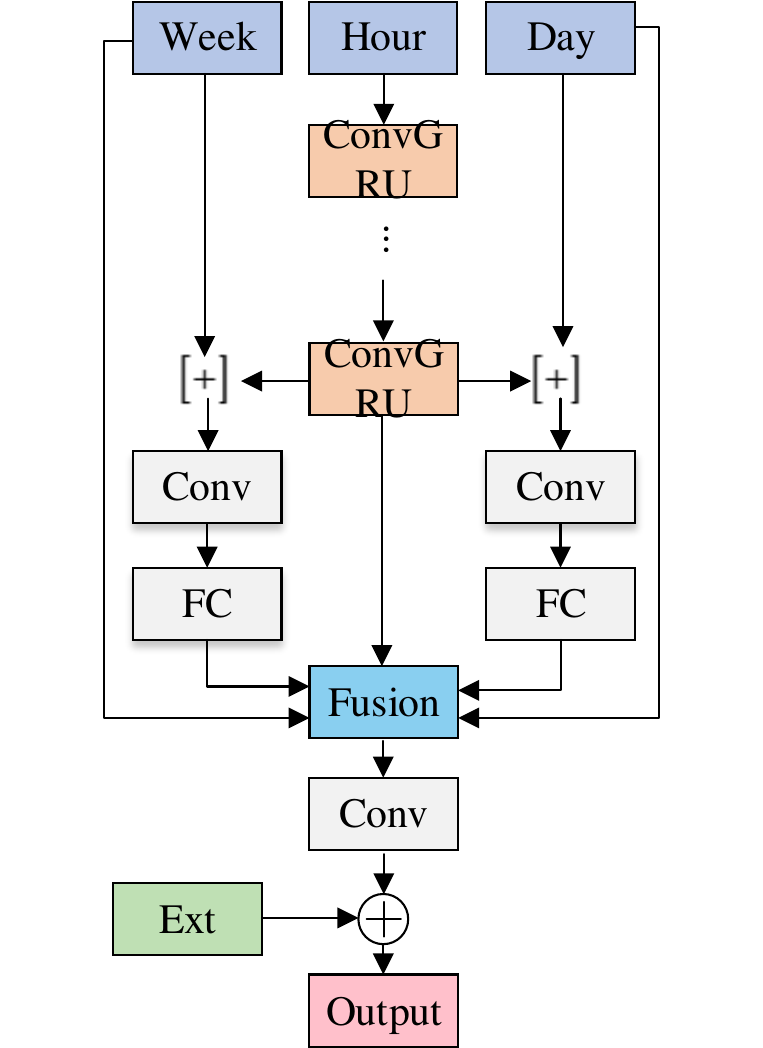}}
    \quad
    \subfloat[STAR]{\label{fig:model3}
    \includegraphics[height=1.9in]{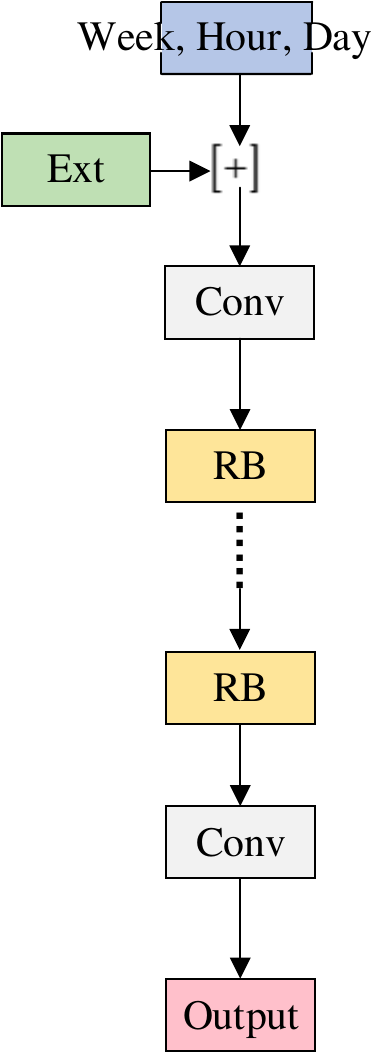}}\\
    \caption{Simplified structures of recent DL models. \textit{(RB means a residual block, including two weight layers. Conv, FC, and GRU represent convolution layer, fully-connected layer and gate recurrent units respectively. In all four cases, the green refers to the process of encoding external factors, the light blue refers to the fusion component, and the outputs and inputs with pink and purple color are supervised. $[+]$ and $\oplus $ are the concatenate and element-wise addition respectively)}}
    \label{fig:models}
\end{figure}

Since the grid-based method is proposed, several DL models, such as ST-ResNet \cite{zhang2017deep}, MST3D \cite{chen2018exploiting} and PCRN \cite{zonoozi2018periodic} have also been proposed. These models take multiple components to learn temporal correlation and then fuse the external influence to derive the final result. Fig.~\ref{fig:models} illustrates these models in simplified structures, where the activation functions and batch normalization \cite{ioffe2015batch} are omitted for clarity.

In ST-ResNet, the historical data are separated into three same fully-convolutional residual networks to capture temporal dependencies, then parametric-matrix fusion (PM fusion) is proposed to adjust the degrees affected by three components. In terms of spatial dependencies, the main idea of ST-ResNet is to use the convolutional neural network (CNN) with the residual block (RB) sequence to training deep networks.

Differing from ST-ResNet, MST3D employs 3D convolutional layers with downsampling instead of 2D convolutional layers, and residual learning is not adopted, so the input sequence of MST3D is longer than ST-ResNet to capture temporal dependencies in each component with only three weight layers. 

The motivation for PCRN is to observe recurring periodic patterns in ST data. Therefore, PCRN adopts convolutional GRUs to learning ST representation, which is dynamically updated from multiple periodic patterns through softmax operation and PM fusion.

\section{Proposed Method}
\subsection{Convert Data and Construct Training Instances}
In our method, the process of convert data and construct training instances is shown in Fig.~\ref{fig:star}. In order to preserve the ST dependencies of trajectory data, we also adopt the grid-based method to split the target city into $I\times J$ regions and count the total number of mobility events passing each region. By this way, the trajectory data is transformed into the image-like observations.
\begin{figure}[!t]
\removelatexerror
\begin{algorithm}[H]
\label{alg:algorithm1}
\caption{selectKeyframes}
$\mathcal{Q}\gets \emptyset$\  \tcp*[r]{initialize a FIFO queue}
\For{$i\in[1,l_c]$}
{
    $\mathcal{Q}\gets \bm{X}_{t-i}$ 
    \tcp*[r]{append closeness}
}
\For{$i\in[1,l_p]$}
{
    \For{$r\in[0,l_r]$}
    {
        $\mathcal{Q}\gets \bm{X}_{t-i\times p-r}$         \tcp*[r]{append period}
    }
}
\For{$i\in[1,l_q]$}
{
    \For{$r\in[0,l_r]$}
    {
        $\mathcal{Q}\gets \bm{X}_{t-i\times q-r}$ 
        \tcp*[r]{append trend}
    }
}
\Return{$\mathcal{Q}$}
\end{algorithm}
\end{figure}

The numerous observations $\{\bm{X}_1,\bm{X}_2,...,\bm{X}_{t-1}\}$ are generated by the above process, but we cannot feed them all into the model, for that will make the whole training processing non-trivial. Therefore, we leverage temporal closeness, period and trend to select keyframes for modeling as shown in the left side of Fig.~\ref{fig:star}. Formally, assuming $l_c,l_p$ and $l_q $ are lengths of closeness, period, trend fragment respectively, our method uses Algorithm~\ref{alg:algorithm1} to select keyframes queue $\mathcal{Q}$, where $p$ and $q$ refer to period and trend span respectively. $r$ is the length of the sub-fragment. We add a hyperparameter $r$ to extend an interval to multiple intervals to increase input information. For example, if we choose 30 minutes as a time interval, parameters $l_c,l_p,l_q $ and $r$ are set to 2, 1, 1 and 1, respectively. $p$ and $q$ are set to 48 (one-day) and 336 (one-week). $[\bm{X}_{t-1},\bm{X}_{t-2}]$,$[\bm{X}_{t-48}]$ and $[\bm{X}_{t-336}]$ frames are selected in ST-ResNet, while we select $[\bm{X}_{t-1},\bm{X}_{t-2},\bm{X}_{t-48},\bm{X}_{t-49},\bm{X}_{t-336},\bm{X}_{t-337}]$ as keyframes. Additional $\bm{X}_{t-49}$ and $\bm{X}_{t-337}$ are relatively associated with $\bm{X}_{t-48}$ and $\bm{X}_{t-336}$ in closeness dependents, while they also share similarities with $\bm{X}_{t-2}$ in period and trend dependents. Then we stack queue $\mathcal{Q}$ into a tensor $\bm{I}_t\in \mathbb{R}^{(6\times 2)\times I\times J}$ as the input of the model, where 2 refers to two types of flows. The reasons for using $\bm{I}_t$ are two-fold. First, the convolution kernel in the channel dimension can take relevance of each keyframes pattern into account. This simple way shows better performance than the sophisticated fusion methods. Second, the number of parameters is reduced significantly by learning a single tensor $\bm{I}_t$ for aggregated representations.

\subsection{Convolution Operation}


CNN is effective in exploiting spatial dependencies by using the local connection \cite{lecun1998gradient}, which has more advantages than other types of neural networks in ST prediction task \cite{zhang2017deep}. For this reason, we mainly use convolution layers in our framework. Formally, if we consider the channel dimension $c$ the $(j,k)$ element $\bm{T}^l_{i,j,k}$ of the output of $l$-th convolution layer, the $i$-th channel is generated by convolution kernel $\bm{K}_{i,l,m,n}$ of $m\times n$ size, which is calculated by

\begin{equation}
    \bm{T}^l_{i,j,k}=\sum _c \sum_m \sum_n \bm{T}^{l-1}_{c,j+m,k+n}\bm{K}_{i,c,m,n}
\end{equation}
note that, $\sum _c$ refers to the summation of convolution operation in channel dimension, which takes into account the effects of all frames on learning temporal dependencies.

The discussion has arisen in previous work that whether 2D or 3D convolution is more efficient in ST tasks \cite{chen2018exploiting,xie2018rethinking,tran2018closer}, while this is an issue concerning the selection and trade-off based on data features, the amount of data, time cost and model capacity. Suppose the input tensor shape is $L\times I\times J\times 2$, where $L$ is the number of frames in the keyframes queue, $I$ and $J$ are the frame height and width, and 2 refers to two types of flows. 2D convolution treats the $L$ frames analogously to channels. Thus, we can think of 4D tensor into a 3D tensor of size $I\times J\times 2L$. One problem with this operation is ignoring the temporal ordering, which is very necessary for temporal data with a continuous period, especially for an extended period, like action recognition \cite{tran2018closer}. In this work, one of the main reasons that we employ 2D convolution is to concatenate frames of recent, near and distant into a tensor $\bm{L}_t$, which do not have strong temporality. Besides, experiments in \cite{zhang2018predicting} show that long-range periods and trend are hard to model or helpless. In the special implementation, we only choose very few keyframes. Therefore, using 2D convolution almost does not have the above problem, and the experiment shows 2D convolution performs better than 3D convolution because the convolution kernels consider all frames (channel dimensions) at each convolution operation.

\subsection{Convolutional Residual Blocks}
As we know, the scale of a city is generally large. To capture the spatial attributes $\bm{S}$ of a city with the large dimensions of $I$ and $J$, we expect more convolution layers to capture farther dependencies, even citywide spatial dependencies. In order to accomplish this, we employ residual learning \cite{he2016deep} for training deep network.

In this work, each RB consists of $l_w$ convolution layers with a rectifier linear unit (ReLU) \cite{nair2010rectified} before each convolution layer, which is "pre-activation" structure \cite{he2016identity}. The output of the $l$-th residual block is given by

\begin{equation}
    \bm{T}^l=\mathcal{F}(\bm{T}^{l-1};\theta^l)+\bm{T}^{l-1}
\end{equation}
where $\bm{T}^{l-1}$ and $\bm{T}^l$ are the input and output of the $l$-th RB. As shown in the yellow component of Fig.~\ref{fig:star}, $\mathcal{F}(;\theta^l)$ denotes the residual function, which implements the composition of two convolutions parameterized by weights and the application of the ReLU. 

\subsection{STAR}

\begin{figure}[t!]
\centerline{\includegraphics[width=\columnwidth]{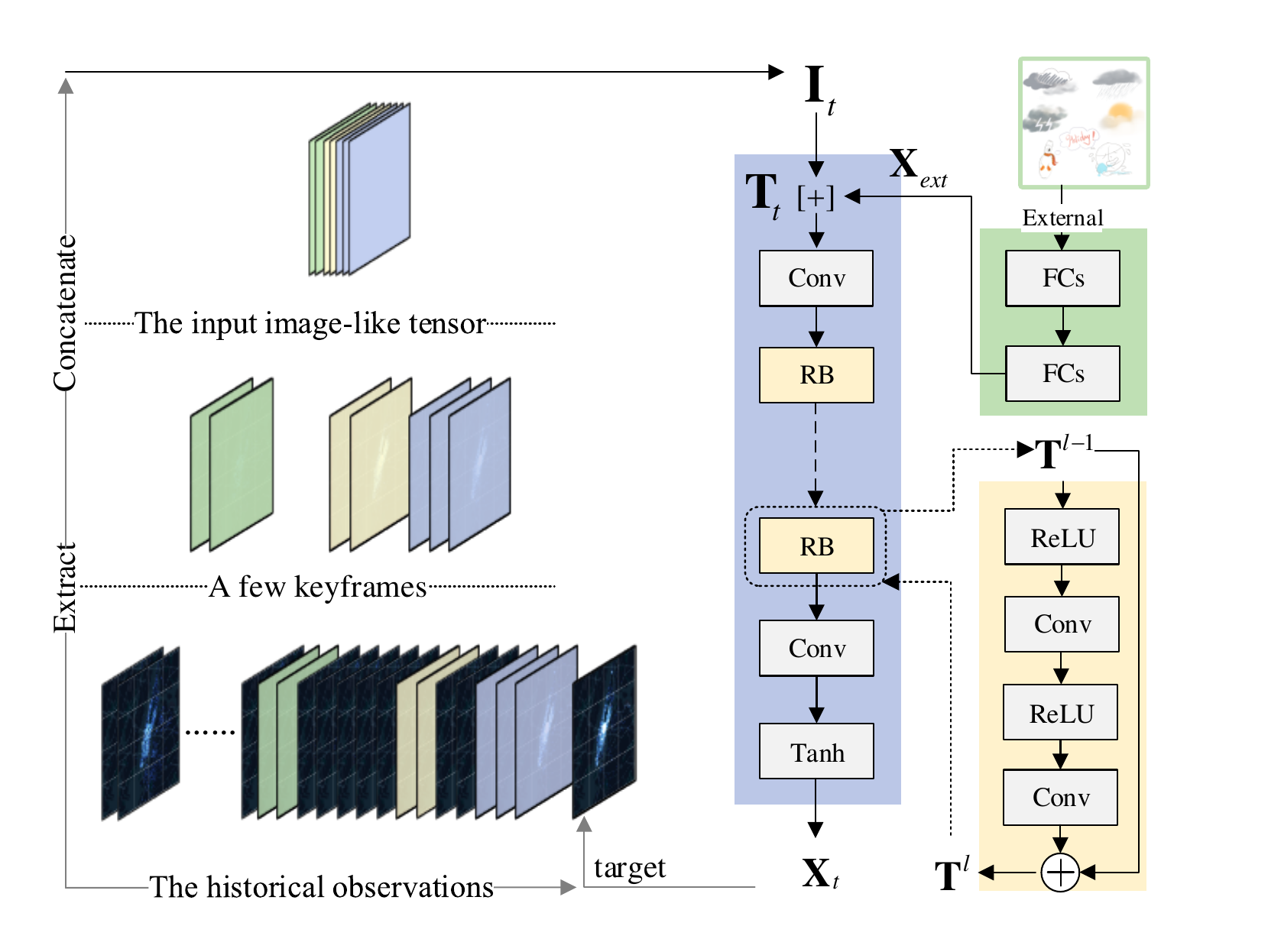}}
\caption{Structures of STAR. \textit{(The left side of the figure is the pipeline of construct input data $\bm{I}_t$, which contains two steps to extract and concatenate keyframes. The green component is the external component to generate  $\bm{X}_{ext}$, $[+]$ means concatenate $\bm{I}_t$ and $\bm{X}_{ext}$ into $\bm{T}_{t}$. The yellow component is the detail structures of residual function, and the purple component is the main network)}}
\label{fig:star}
\end{figure}

We use the fully-convolutional residual network, which is similar to ST-ResNet. The most significant difference between our method and ST-ResNet is that STAR only uses a single network and the way of the input data constructed changes as well. Both of our method and ST-ResNet are shown in Fig.~\ref{fig:models}. In detail, our network structure has two components shown in Fig.~\ref{fig:star}. 

The green component is the external component, which adopts two FC layers. The first layer is used to embedding external factors. The last layer is used to map low to high dimensions that have the same shape as $\bm{X}_t$. Then the external component's output $\bm{X}_{ext}$ is concatenated to main input $\bm{I}_t$ with the channel axis, as one tensor $\bm{T}_t$.

The purple component is a convolutional neural network with $L$ RBs. We use $2+L\times l_w$ layers where layers except the first and the last are of the same RB (e.g. 64 kernels of the size $3\times 3\times 64$, where a kernel operates on $3\times 3$ spatial regions across 64 feature maps). The first layer operates on fusion tensor $\bm{T}_t$ to generate feature maps with same dimensions as the next layer. The last layer used for human mobility prediction consists of two kernels of size $3\times 3$. For all layers except the last, ReLU is taken as the activation function. The last layer uses Tanh as the output function, for mapping the output to the range $[-1,1]$.

One problem with using a deep network to predict outputs is that the size of the feature map gets reduced every time convolution operation is applied. In our task, the final output size should be the same as the size of the groundtruth. For this goal, we pad zeros before convolution to keep the shape of all feature maps and output layer.

\section{Experiment}
\subsection{Experiment Settings}
\noindent {\bf Environment.} We conduct experiments on a 64-bit Ubuntu 16.04 computer with an Intel 3.40 GHz and an NVIDIA GTX 1070 GPU. The proposed method is implemented with Python 3.6, Keras 2.1.5 and Tensorflow 1.6.

\noindent {\bf Datasets.} We evaluate our framework on two benchmark datasets as follows. (1) TaxiBJ is the trajectory data of the taxicab GPS data and meteorology data in Beijing in eighteen discrete months. The city of Beijing is divided into $32\times 32$ individual regions, and inflow and outflow in intervals of 30 minutes are counted, resulting in $\bm{X}_t\in \mathbb{R}^{2\times 32 \times 32}$ image-like observations. We choose the last 4 weeks as the testing data. External information includes holidays, weather conditions and temperature. (2) BikeNYC is the trip data, which is taken from New York City (NYC) Bike system in six consecutive months. We split NYC into $16\times 8$ individual regions, using the above method to produce $\bm{X}_t\in \mathbb{R}^{2\times 16 \times 8}$ observations. The last 10 days are chosen as testing data. Detail of datasets description could be found in \cite{zhang2017deep}.

\subsection{Implementation Details}
For comparison, we follow ST-ResNet in pre-processing, modeling and evaluation, and the details are as follows.

\noindent {\bf Pre-Processing.} Min-max normalization method is used to convert the train data by $[-1,1]$ scale and one-hot coding is used to transform metadata. 

\noindent {\bf Modeling.} Adaptive moment estimation (Adam) is adopted as the optimization method with a fixed learning rate and mean squared error (MSE) as the loss function. The learnable parameters are initialized using the default parameter in Keras. For TaxiBJ dataset, we apply 6 RBs, which implements the composition of 2 weight layers. All convolutions except the last layer use 64 filters. For BikeNYC dataset, 2 RBs with the composition of one weight layers are applied. The number of convolutional filters on all convolution layers except the last layer is set to 256, and L2 regularization is applied to the kernel to reduce the over-learning issue. One of the main reasons is that the dimension of spatial and the amount of data are smaller than that of the TaxiBJ dataset. In addition to this, the kernel size in two datasets are set to $3\times 3$, and the hyperparameters $l_c,l_p,l_q$ and $r$ are set to 3, 1, 1 and 2, respectively. $p$ and $q$ are empirically fixed to one-day and one-week, respectively. The batch size is 16. We choose the same amount of data as the test set from the training set as the verification set, which is used to early-stop in our training. Afterwards, we use full training data to retrain the model for 100 epochs.

\noindent {\bf Evaluating.} RMSE is used as the evaluation metric for evaluating the performance of each model,

\begin{equation}
    RMSE=\sqrt{\frac{1}{N}\sum_{t=1}^N(\hat{\bm{X}_t}-\bm{X}_t)^2}
\end{equation}
where $\bm{X}_t$ is the groundtruth, $\hat{\bm{X}_t}$ is the corresponding predicted values, and $N$ is the number of all groundtruth. 

\subsection{Comparison with State-of-the-Art Models}
We compare our model with the best performing variants of the above three DL methods. Experimental setting is kept the same as those previous methods. For a fair comparison, we only use time metadata in the external component (PCRN does not consider other factors). Table 1 shows the performance of our model on TaxiBJ and BikeNYC. It is observed that our model provides more significant outperform than the other DL methods on RMSE. Note that, these methods apply the additional fusion methods, but we only take advantage of the summation of convolution operation in channel dimension, which illustrates that the self-learning ability of deep learning is incredible.

\begin{table}[t!]
\caption{Comparison with State-of-the-Art Models}
\begin{center}
\begin{tabular}{|l|c|c|}
\hline
\textbf{Model/}&\multicolumn{2}{|c|}{\textbf{RMSE}} \\
\cline{2-3} 
\textbf{Method} & \textbf{\textit{TaxiBJ}}& \textbf{\textit{BikeNYC}}\\
\hline
ST-ResNet   & 16.88 & 6.33      \\
\hline
MST3D       & 16.05 & 5.81     \\
\hline
PCRN        & 15.85 & -     \\
\hline
STAR        & \bf{15.59} & \bf{5.75}     \\
\hline
\end{tabular} 
\label{tab1}
\end{center}
\end{table}

\begin{figure}[!t]
    \centering
    \subfloat[inflow]{\label{fig:in}
    \includegraphics[height=0.9in]{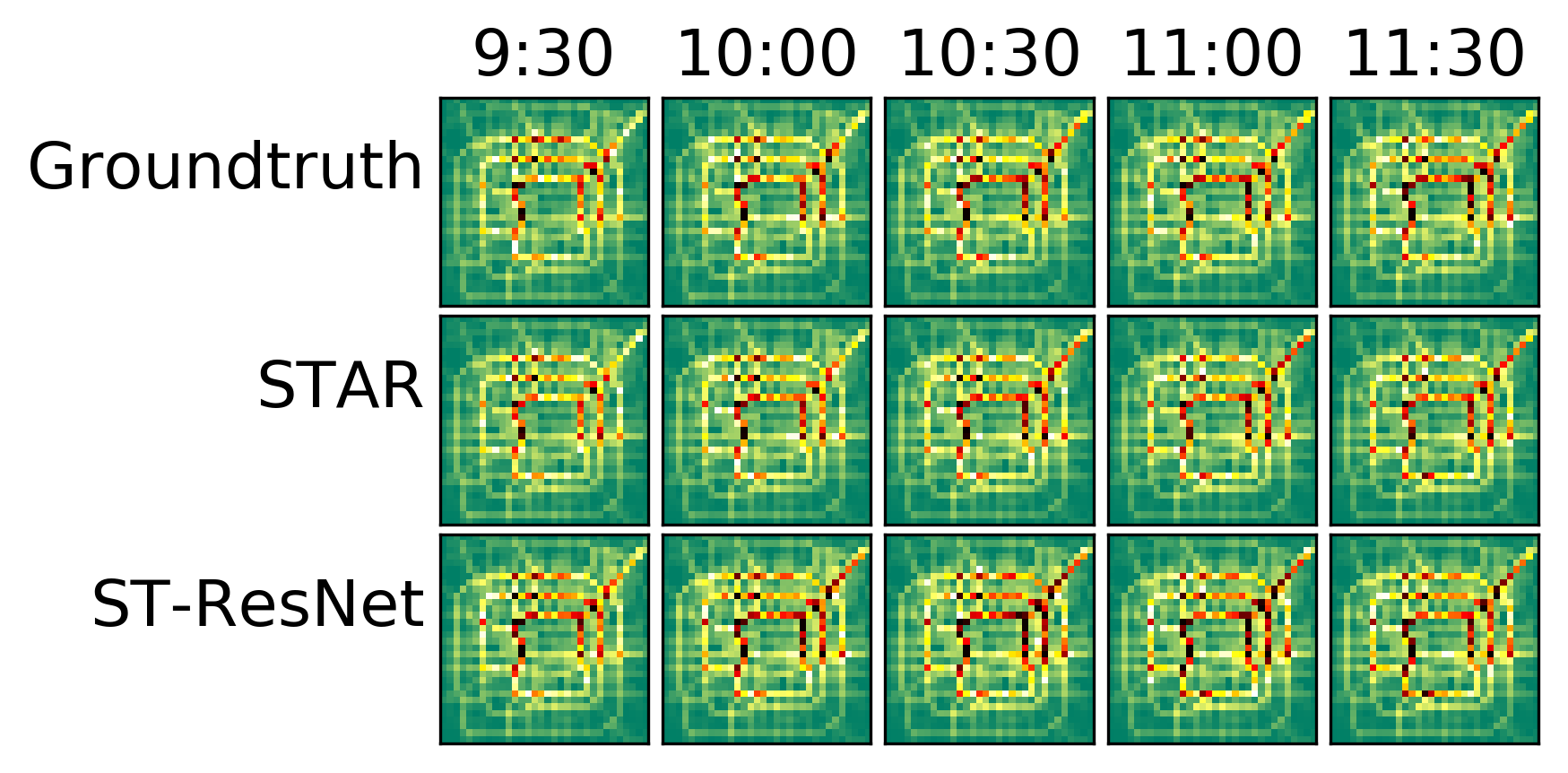}}
    \subfloat[outflow]{\label{fig:out}
    \includegraphics[height=0.9in]{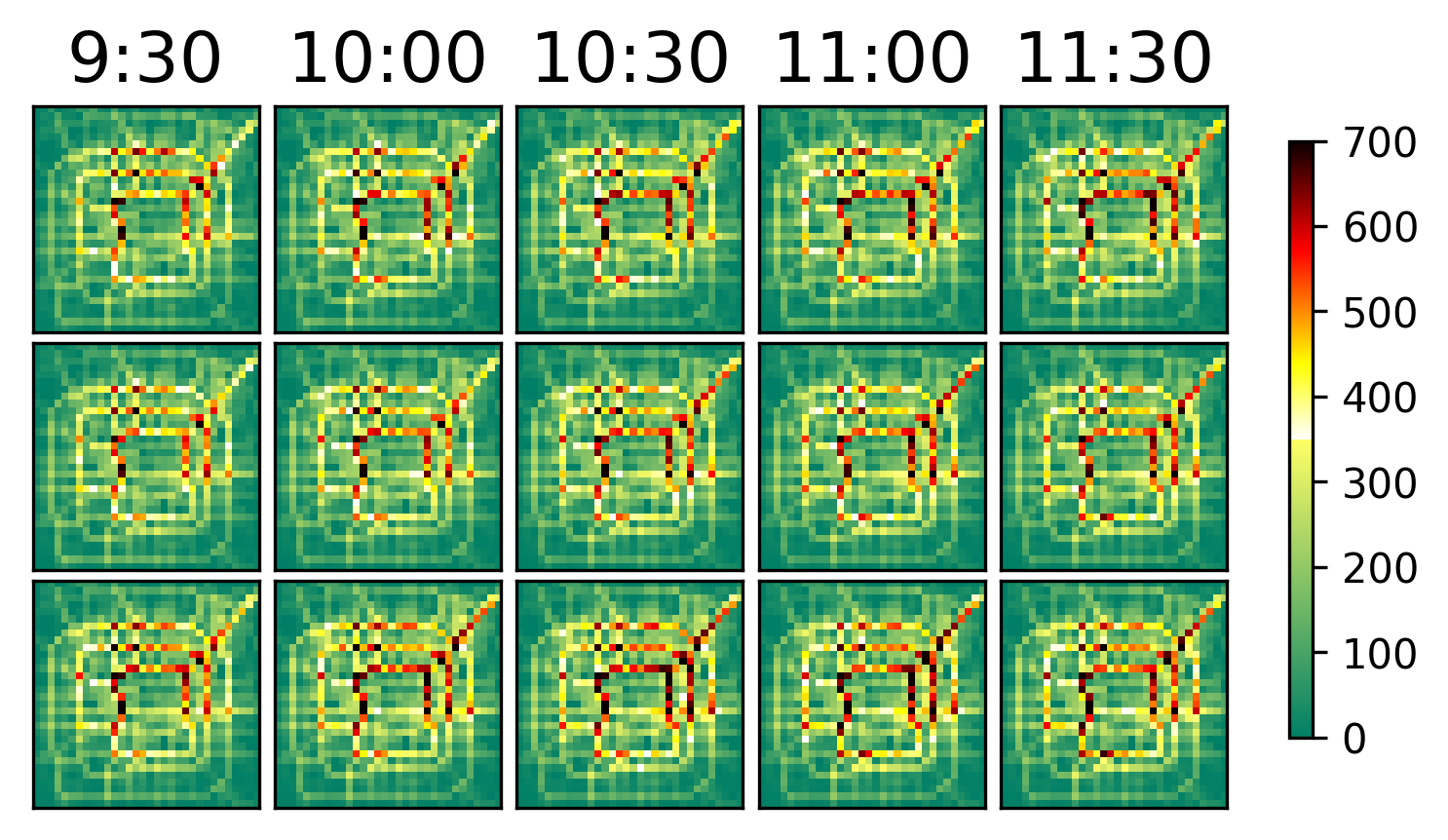}}\\
    \caption{Samples of 5-step ahead prediction on TaxiBJ. \textit{(Images from top to bottom row are: groundtruth; STAR predictions; ST-ResNet predictions)}}
    \label{fig:flow}
\end{figure}

\begin{figure}[t!]
\centerline{\includegraphics[width=3.4in]{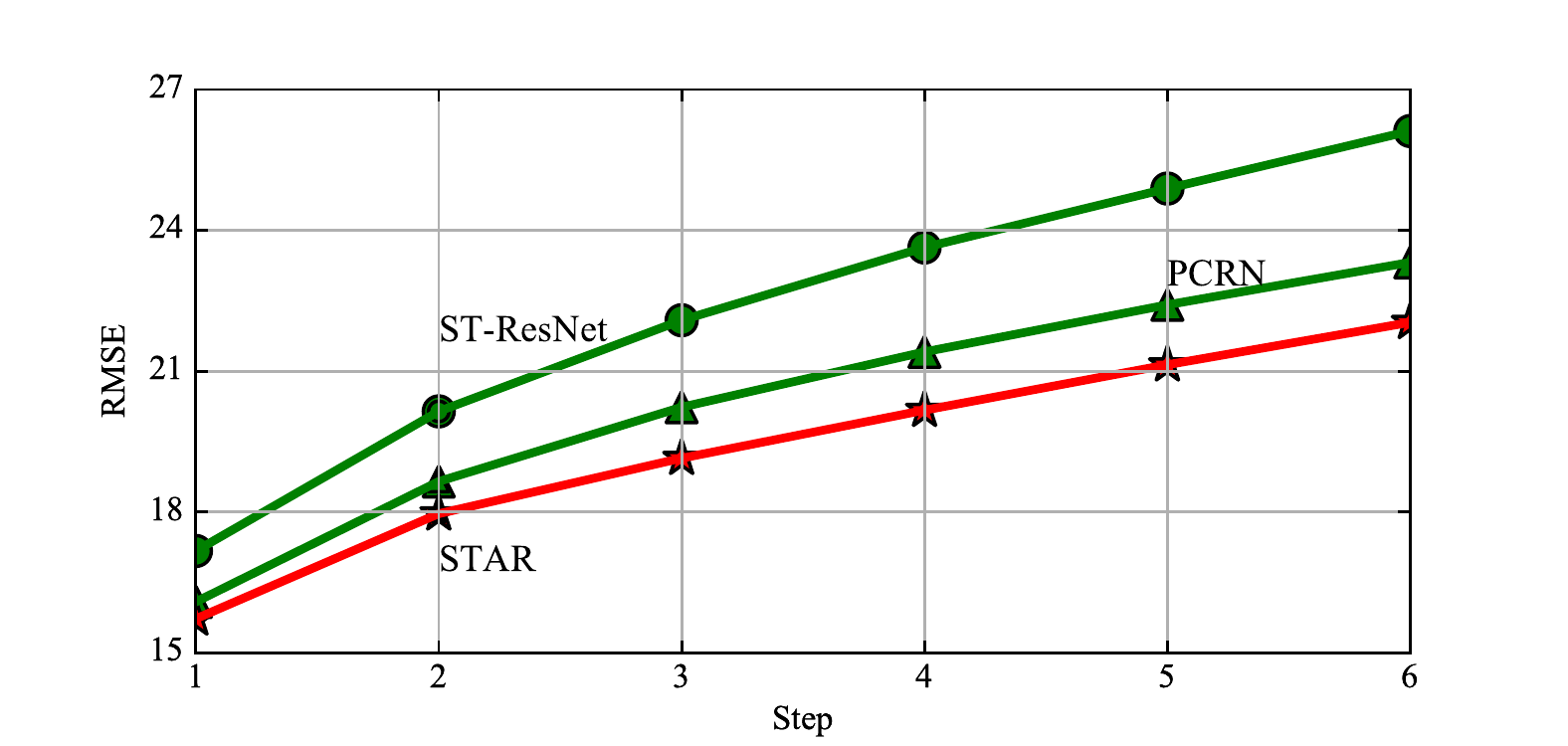}}
\caption{Evaluation of multi-step ahead prediction. \textit{(Redline is our model, which achieves lowest RMSE overall steps than state-of-the-art methods)}}
\label{fig:mulit}
\end{figure}

Moreover, we compare with the best performing variant of state-of-the-art models for predicting the next 6 steps. Multi-step ahead prediction refers to the use of historical observations and the recent predicted results to predict flow in subsequent time intervals. Proceed as follows, learned and saved model first, running the model for a fixed number of steps by appending the output $\bm{X}_t$ for $\bm{T}_t$ to the input $\bm{T}_{t+1}$ for $\bm{X}_{t+1}$ in every step, where $t$ refers to the current step and $t+1$ means the next step. We visualize the 5-step ahead prediction in Fig.~\ref{fig:flow} and observe that STAR usually gives more accurate predictions. For instance, the hot red regions denote massive crowd flows, where possibly the traffic jam happens. For these areas, ST-ResNet's estimates are higher than groundtruth, indicating a relatively high false positive rate. Furthermore, the global results in Fig.~\ref{fig:mulit} show STAR achieves lowest RMSE overall steps.

\subsection{Different STAR Variants }

\begin{table}[t!]
\caption{Comparison of different STAR variants}
\begin{center}
\begin{tabular}{|l|c|}
\hline
\textbf{Model/Method}&{\textbf{RMSE}} \\
\hline
STAR-3D         & 16.01      \\
\hline
STAR-END        & 15.87      \\
\hline
STAR-ST311      & 16.31    \\
\hline
STAR            & \bf{15.59}     \\
\hline
\end{tabular}
\label{tab2}
\end{center}
\end{table}
To illustrate the effectiveness of our method, we studied the performance of 3 different STAR variants on TaxiBJ, namely, STAR-3D, STAR-END, and STAR-ST311. In particular, STAR-3D applies 3D convolution of kernel size of $2\times 3\times 3$ which uses padding zeros to avoid changing the shape. The last layer uses kernel size of  $8\times 3\times 3$ for output prediction. STAR-END adds the external factors before the output. $[\bm{X}_{t-1},\bm{X}_{t-2},\bm{X}_{t-3},\bm{X}_{t-48},\bm{X}_{t-336}]$ frames are stacked as input tensor in STAR-ST311, which is the same as selected in ST-ResNet. Table 2 lists the performance of these variants. Compared with ours, STAR-3D performs worse than STAR, meaning 3D convolution is not suitable for learning keyframes selected in this paper. STAR-ST311 is worse for the keyframes selected is not representative enough. STAR-END is worse than STAR , and one possible reason is the high-level layer of a deep network has strong “semantic” features, while the external component is a map of low to high dimensions (e.g., 8 to $32\times 32\times 2$ bytes), which lacks enough information to merge in the high-level layer. 

\subsection{Efficiency}

\begin{figure}[!t]
    \centering
    \subfloat[]{\label{fig:time}
    \includegraphics[width=1.6in]{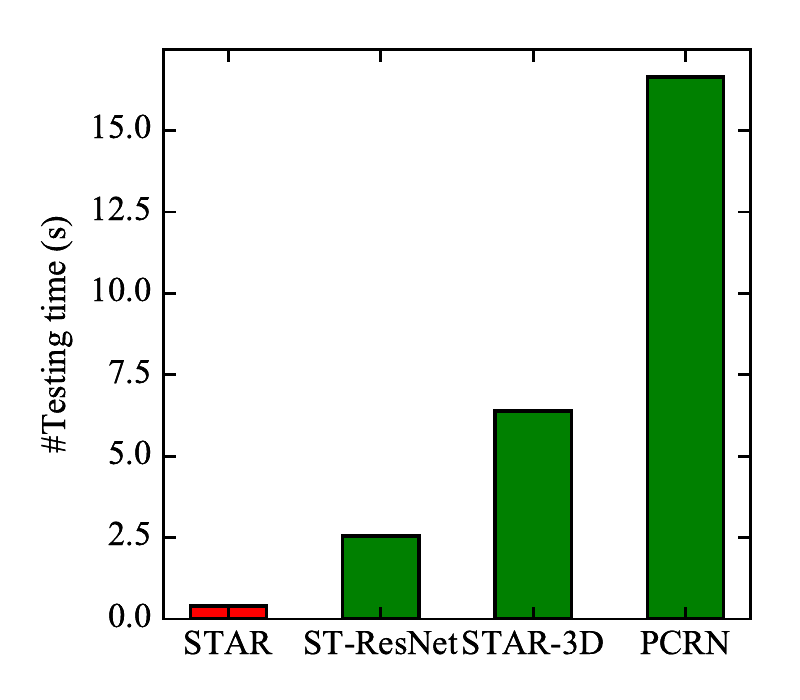}}
    \subfloat[]{\label{fig:parameters}
    \includegraphics[width=1.6in]{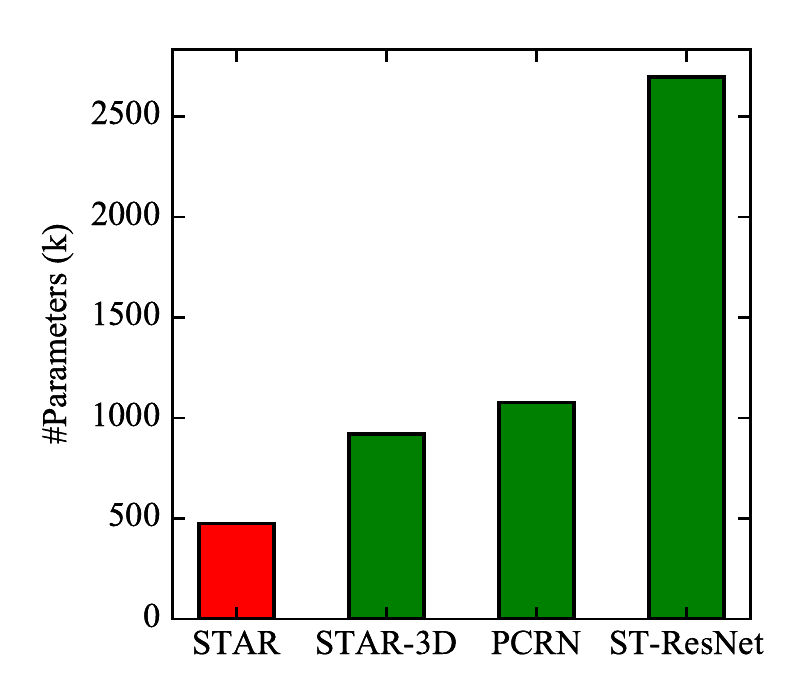}}\\
    \caption{The ranking of recent methods of \protect\subref{fig:time} running time and \protect\subref{fig:parameters} model size. \textit{(STAR displays the fastest running time and the fewest parameters)}}
    \label{fig:time}
\end{figure}

As stated in the introduction, the utmost important factors of citywide human mobility prediction lie in high efficiency. Fig.~\ref{fig:time} shows the ranking of recent DL methods of running time and model size, respectively. Note that, testing time usually is positively correlated with training time, especially for the framework with the recurrent structure. In general, 3D- convolution-based method and recurrent structure usually have fewer parameters but more time cost, while 2D convolution has a distinct advantage in operation speed. Specifically, our model displays the fastest speed of 0.4 (s) and the fewest parameters of 476.2 (k), which are 41.6x faster than PCRN and 5.7x fewer than ST-ResNet respectively, which proves the effective methods are usually simple and general in practical problems.

\section{Conclusion and Future Work}
The paper proposes a simplified fully-convolutional residual network for citywide human mobility forecasting, which processes the spatial-temporal data as the multi-channel image. The experimental results show that our framework achieves state-of-the-art performance on two types of mobility event in Beijing and NYC. Possible future work might include exploring multi-source data or multi-task joint mining in an overall model. The code can be found in \url{https://github.com/hongnianwang/STAR}

\section*{Acknowledgment}
This work was supported by the National Science Foundation of China under
Grant Nos. 61403266 and 61403196, Chinese overseas returnees science and technology activities project funding of Ministry of Human Resources and Social Security.

\bstctlcite{IEEEexample:BSTcontrol}

\bibliographystyle{IEEEtran}
\bibliography{mdm}

\begin{thebibliography}{10}
\providecommand{\url}[1]{#1}
\csname url@samestyle\endcsname
\providecommand{\newblock}{\relax}
\providecommand{\bibinfo}[2]{#2}
\providecommand{\BIBentrySTDinterwordspacing}{\spaceskip=0pt\relax}
\providecommand{\BIBentryALTinterwordstretchfactor}{4}
\providecommand{\BIBentryALTinterwordspacing}{\spaceskip=\fontdimen2\font plus
\BIBentryALTinterwordstretchfactor\fontdimen3\font minus
  \fontdimen4\font\relax}
\providecommand{\BIBforeignlanguage}[2]{{%
\expandafter\ifx\csname l@#1\endcsname\relax
\typeout{** WARNING: IEEEtran.bst: No hyphenation pattern has been}%
\typeout{** loaded for the language `#1'. Using the pattern for}%
\typeout{** the default language instead.}%
\else
\language=\csname l@#1\endcsname
\fi
#2}}
\providecommand{\BIBdecl}{\relax}
\BIBdecl

\bibitem{zheng2014urban}
Y.~Zheng, L.~Capra, O.~Wolfson, and H.~Yang, ``Urban computing: concepts,
  methodologies, and applications,'' \emph{ACM Transactions on Intelligent
  Systems and Technology}, vol.~5, no.~3, pp. 38:1--38:55, Sept 2014.

\bibitem{zhang2017deep}
J.~Zhang, Y.~Zheng, and D.~Qi, ``Deep spatio-temporal residual networks for
  citywide crowd flows prediction.'' in \emph{AAAI Conference on Artificial
  Intelligence}, Feb 2017, pp. 1655--1661.

\bibitem{liu2018urban}
Z.~Liu, Z.~Li, K.~Wu, and M.~Li, ``Urban traffic prediction from mobility data
  using deep learning,'' \emph{IEEE Network}, vol.~32, no.~4, pp. 40--46, July
  2018.

\bibitem{lv2015traffic}
Y.~Lv, Y.~Duan, W.~Kang, Z.~Li, F.-Y. Wang \emph{et~al.}, ``Traffic flow
  prediction with big data: A deep learning approach.'' \emph{IEEE Transactions
  on Intelligent Transportation Systems}, vol.~16, no.~2, pp. 865--873, April
  2015.

\bibitem{zheng2015trajectory}
Y.~Zheng, ``Trajectory data mining: an overview,'' \emph{ACM Transactions on
  Intelligent Systems and Technology}, vol.~6, no.~3, pp. 29:1--29:41, May
  2015.

\bibitem{lecun2015deep}
Y.~LeCun, Y.~Bengio, and G.~Hinton, ``Deep learning,'' \emph{Nature}, vol. 521,
  no. 7553, pp. 436--444, May 2015.

\bibitem{yuan2018hetero}
Z.~Yuan, X.~Zhou, and T.~Yang, ``Hetero-convlstm: A deep learning approach to
  traffic accident prediction on heterogeneous spatio-temporal data,'' in
  \emph{ACM SIGKDD International Conference on Knowledge Discovery \& Data
  Mining}, Aug 2018, pp. 984--992.

\bibitem{chen2018exploiting}
C.~Chen, K.~Li, S.~G. Teo, G.~Chen, X.~Zou, X.~Yang, R.~C. Vijay, J.~Feng, and
  Z.~Zeng, ``Exploiting spatio-temporal correlations with multiple 3d
  convolutional neural networks for citywide vehicle flow prediction,'' in
  \emph{IEEE International Conference on Data Mining}, Nov 2018, pp. 893--898.

\bibitem{zonoozi2018periodic}
A.~Zonoozi, J.-j. Kim, X.-L. Li, and G.~Cong, ``Periodic-crn: A convolutional
  recurrent model for crowd density prediction with recurring periodic
  patterns.'' in \emph{International Joint Conference on Artificial
  Intelligence}, 2018, pp. 3732--3738.

\bibitem{ioffe2015batch}
S.~Ioffe and C.~Szegedy, ``Batch normalization: Accelerating deep network
  training by reducing internal covariate shift,'' \emph{arXiv preprint
  arXiv:1502.03167}, 2015.

\bibitem{lecun1998gradient}
Y.~LeCun, L.~Bottou, Y.~Bengio, and P.~Haffner, ``Gradient-based learning
  applied to document recognition,'' \emph{Proceedings of the IEEE}, vol.~86,
  no.~11, pp. 2278--2324, Nov 1998.

\bibitem{xie2018rethinking}
S.~Xie, C.~Sun, J.~Huang, Z.~Tu, and K.~Murphy, ``Rethinking spatiotemporal
  feature learning: Speed-accuracy trade-offs in video classification,'' in
  \emph{European Conference on Computer Vision}, Oct 2018, pp. 305--321.

\bibitem{tran2018closer}
D.~Tran, H.~Wang, L.~Torresani, J.~Ray, Y.~LeCun, and M.~Paluri, ``A closer
  look at spatiotemporal convolutions for action recognition,'' in
  \emph{IEEE/CVF Conference on Computer Vision and Pattern Recognition}, June
  2018, pp. 6450--6459.

\bibitem{zhang2018predicting}
J.~Zhang, Y.~Zheng, D.~Qi, R.~Li, X.~Yi, and T.~Li, ``Predicting citywide crowd
  flows using deep spatio-temporal residual networks,'' \emph{Artificial
  Intelligence}, vol. 259, pp. 147--166, June 2018.

\bibitem{he2016deep}
K.~He, X.~Zhang, S.~Ren, and J.~Sun, ``Deep residual learning for image
  recognition,'' in \emph{IEEE Conference on Computer Vision and Pattern
  Recognition}, June 2016, pp. 770--778.

\bibitem{nair2010rectified}
V.~Nair and G.~E. Hinton, ``Rectified linear units improve restricted boltzmann
  machines,'' in \emph{International conference on machine learning}, June
  2010, pp. 807--814.

\bibitem{he2016identity}
K.~He, X.~Zhang, S.~Ren, and J.~Sun, ``Identity mappings in deep residual
  networks,'' in \emph{European conference on computer vision}, Sept 2016, pp.
  630--645.

\end{thebibliography}

\end{document}